\pdfoutput=1

\documentclass[11pt]{article}

\usepackage[preprint]{acl}

\usepackage{times}
\usepackage{latexsym}

\usepackage[T1]{fontenc}

\usepackage[utf8]{inputenc}

\usepackage{microtype}

\usepackage{inconsolata}

\usepackage{graphicx}

\usepackage{amssymb}
\usepackage{amsmath}
\newcommand{\para}[1]{\noindent \textbf{#1}}

\newcommand{\precisionsubs}{{Prec}}

\newcommand{\avgsimmsubs}{Rec}

\newcommand{\fsubs}{{F1}}

\newcommand{\ours}{OVFact} 

\newcommand{\precision}{OVFact$_{\precisionsubs{}}$}

\newcommand{\avgsim}{OVFact$_{\avgsimmsubs}$}

\newcommand{\fours}{OVFact$_{\fsubs{}}$}

\usepackage{cleveref}
\usepackage{booktabs}
\usepackage{pgfplots}
\usepackage{pgfplotstable}
\usepackage{multirow}

\newlength{\fHeight}
\newcommand{\faEnvelopeO}{\includegraphics[height=\fHeight]{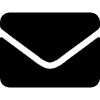}}

\title{OVFact: Measuring and Improving Open-Vocabulary Factuality \\ for Long Caption Models}

\author{
 \textbf{Monika Wysoczańska\textsuperscript{1,2}\thanks{Work done as a Student Researcher at Google DeepMind}} \hspace{0.1em}   
 \textbf{Shyamal Buch\textsuperscript{1}} \hspace{0.1em} 
 \textbf{Anurag Arnab\textsuperscript{1}} \hspace{0.1em} 
 \textbf{Cordelia Schmid\textsuperscript{1}} 
\\\\
 \textsuperscript{1}Google DeepMind\quad
 \textsuperscript{2}Warsaw University of Technology
 \\
 \small{
   \faEnvelopeO~: \href{monika.wysoczanska.dokt@pw.edu.pl}{monika.wysoczanska.dokt@pw.edu.pl}
   }
}

\begin{document}
\maketitle
\begin{abstract}
Large vision-language models (VLMs) often struggle to generate long and \textit{factual} captions. However, traditional measures for hallucination and factuality are not well suited for evaluating longer, more diverse captions and in settings where ground-truth human-annotated captions are unavailable. We introduce \ours{}, a novel method for measuring caption factuality of long captions that leverages open-vocabulary visual grounding and tool-based verification without depending on human annotations. Our method improves agreement with human judgments and captures both caption descriptiveness (recall) and factual precision in the same metric. Furthermore, unlike previous metrics, our reference-free method design enables new applications towards factuality-based data filtering. We observe models trained on an \ours{}-filtered (2.5-5x less) subset of a large-scale, noisy (VLM-generated) pretraining set meaningfully improve factuality precision without sacrificing caption descriptiveness across a range of downstream long caption benchmarks.
\end{abstract}

\section{Introduction}

\begin{figure}[t!]
    \vspace{-2\baselineskip}
    \centering
    \includegraphics[width=\linewidth, trim={0em, 0em, 0em, 0em}, clip]{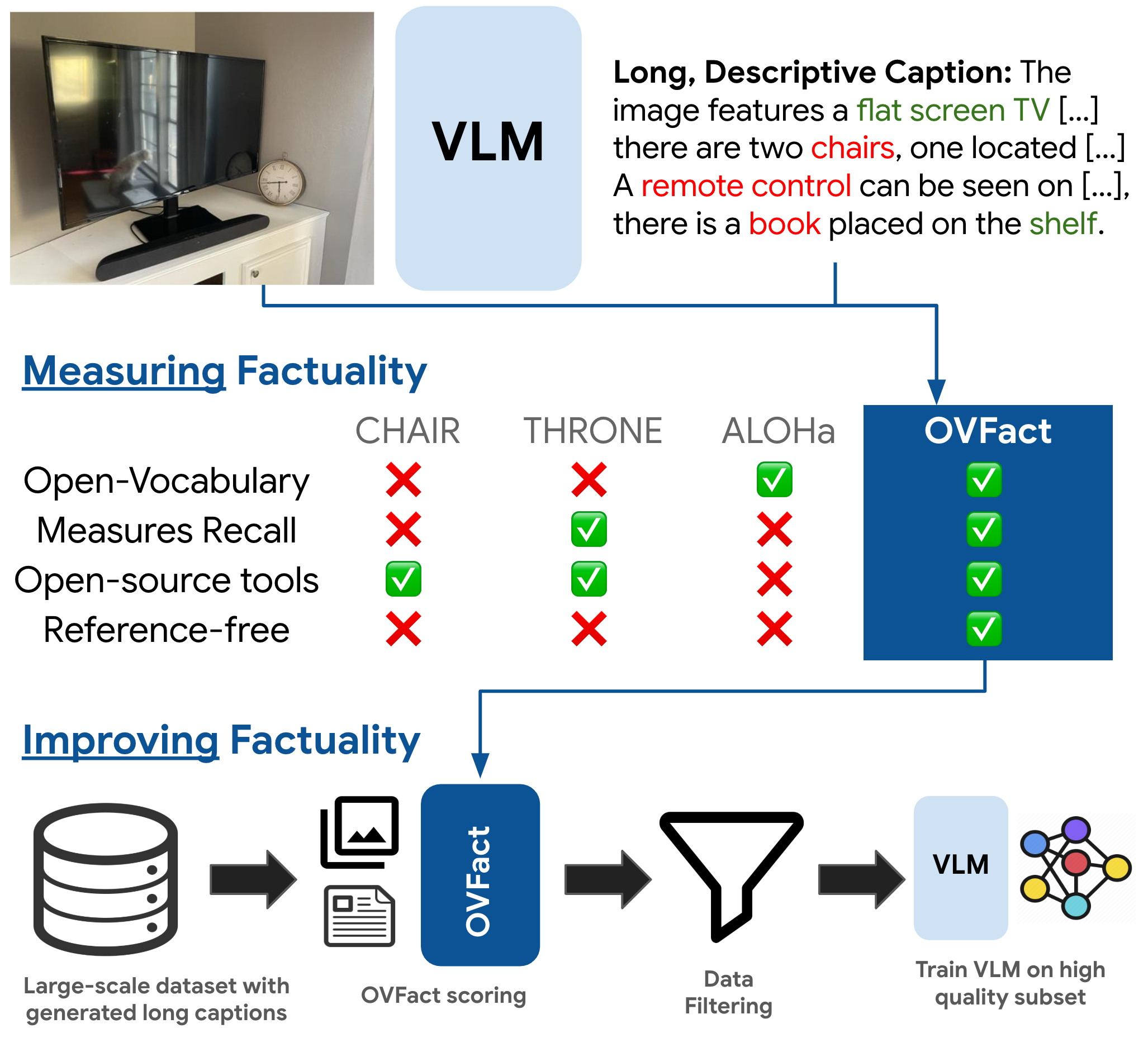} %
    \caption{\textbf{Measuring and improving factuality.} We propose OVFact, a method for measuring and improving \textit{open-vocabulary factuality} of vision-language models (VLMs) in long captioning. (\textbf{top}) Unlike prior work tailored for short captions or specific datasets, 
    \ours{} is \textit{reference-free} and does not require annotated ground-truth.
    (\textbf{bottom}) Our flexible design enables a new factuality-based data filtering application, where models trained on \ours{}-filtered datasets show improved factuality on downstream benchmarks with 2.5-5x less (higher-quality) training data and without compromising caption descriptiveness.
    }
    \vspace{-1.0\baselineskip}
    \label{fig:teaser}
\end{figure}

Large vision-language models (VLMs) are fundamental to a range of grounded language understanding applications, including multimodal AI assistants and tools \cite{achiam2023gpt,georgiev2024gemini}. These models have grown in capability from generating short sentence descriptions \cite{socher2014grounded,karpathy2015deep} to long paragraphs \cite{beyer2024paligemma,steiner2024paligemma,chen2023sharegpt4v}. However, long caption models struggle to maintain \textit{factuality}\footnote{Here, we specifically focus on \textit{object-level} caption factuality: whether noun phrases in VLM generated text are accurate to the original visual input, and vice-versa whether key objects are reflected in the description.}
over these long descriptions \cite{Kaul2024THRONEAO}, including hallucinations of objects that are not present in the input image. As such, there is a crucial need to both measure and improve this limitation.

Prior work for \textit{measuring} factuality for VLMs has been promising but limited. Question-answering approaches \cite{li2023evaluating, jiang2024haleval, huang-etal-2024-lvlms, Guan_2024_CVPR} are indirect, and do not assess if a \textit{particular} output from a model is factual or not \cite{Kaul2024THRONEAO}, as is important in safety-critical settings \cite{bommasani2021opportunities}. On the other hand, metrics for directly assessing caption text \cite{objectHallucination,Kaul2024THRONEAO,petryk2024aloha,ben-kish-etal-2024-mitigating, qiu2024valor} have traditionally been tailored for short captions or specific datasets with limited vocabularies (e.g., MS-COCO \cite{lin2014microsoft}).
 Notably, many of these methods shown in Fig.~\ref{fig:teaser} also rely on ground-truth human annotations, which means they do not work well when such references are unavailable \cite{petryk2024aloha}, as is often the case when VLMs are deployed at scale.
Finally, these metrics often do not capture ``recall'', the coverage of detailed objects in the caption \cite{objectHallucination,petryk2024aloha}. This leads to a conflicting incentive for long caption models to output short, conservative text with fewer objects, reducing the risk of a mistake but sacrificing important details \cite{xue2024xgen}.
The simultaneous lack of these attributes in prior metric designs, highlighted in Fig.~\ref{fig:teaser}, also inhibits integrations with techniques that potentially \textit{improve} factuality in long caption models. For example, outputs from strong VLM models have been used to generate large-scale pretraining datasets for long captions \cite{chen2023sharegpt4v,awadalla2024blip3kale}, but these captions can be prone to factuality errors. Automatically filtering this data to ensure higher quality with previous metrics is not possible since this would require human ground truth.
Thus, there is a critical need for a unified, reference-free, open-vocabulary method for both measuring and improving factuality in long captions and models.

In this work, we make the following contribution: 
(i) we introduce \ours{}, a new method for measuring and improving {\underline{o}}pen-{\underline{v}}ocabulary {\underline{fact}}uality in vision-language models (VLMs) that output long, descriptive captions. Our method leverages open-vocabulary visual grounding and tool-based verification to operate robustly in settings without human annotations. It combines aspects of both precision (minimizing object hallucinations) and recall (ensuring coverage of diverse objects) in the same metric. We validate the combination of these tools with careful analysis and observe that our method improves agreement with human judgments for a range of VLM model outputs, particularly in reference-free settings.
(ii) by addressing the combination of limitations in previous metrics, 
our method design enables new applications towards factuality-based data filtering. We observe that models trained on an OVFact-filtered subset (with 2.5-5x size reduction) of a large-scale, noisy (VLM-generated) pretraining set \cite{chen2023sharegpt4v} meaningfully improve factuality precision without sacrificing caption descriptiveness across a range of downstream benchmarks \cite{onoe2024docci,PontTuset2022locnar,objectHallucination}. These findings are consistent across different model scales and are further validated with human evaluation.

\begin{figure*}[t]
    \vspace{-\baselineskip}
    \centering
    \includegraphics[width=\linewidth]{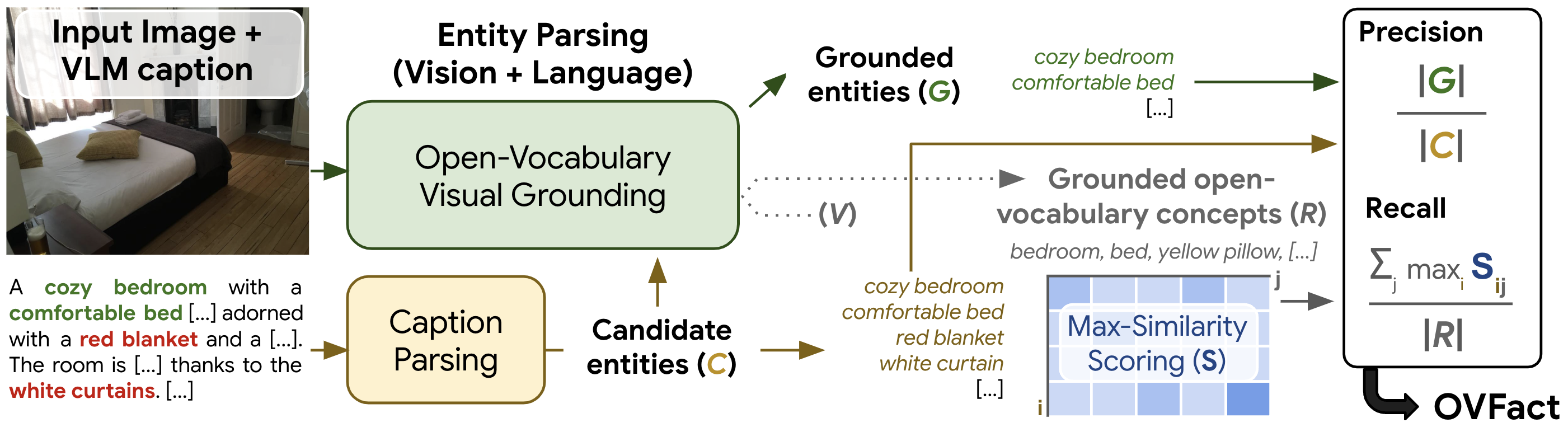}
    \vspace{-1.5\baselineskip}
    \caption{\textbf{\ours{} overview.}
    Our \textit{reference-free} method assesses two aspects of long captioning -- precision and descriptiveness (recall) -- in a unified manner, without requiring ground-truth reference annotations.
    We process a model output caption into a set of candidate entities $\mathcal{C}$, and assess which subset $\mathcal{G}$ are \textit{groundable} in the input image with \textit{open-vocabulary} detection and segmentation tools. Precision is then measured as a ratio of entities remaining (e.g. above``red blanket'' and ``white curtains''detected as hallucinated). To measure descriptiveness, we assess a large open-vocabulary concept set  $\mathcal{V}$ and identify which concepts are grounded in the image $\mathcal{R}$, then measure their recall within the candidate entities from the VLM $\mathcal{C}$ caption using maximum similarity scoring. Unlike prior work \cite{petryk2024aloha,objectHallucination,Kaul2024THRONEAO}, our method can be directly applied to assess factuality in settings where only model-generated caption outputs are available, such as data filtering.}
    \vspace{-\baselineskip}
    \label{fig3:method}
\end{figure*}

\section{Related Work}
\vspace{-0.25\baselineskip}
\label{sec:related}

\paragraph{Object-level factuality in VLMs.}
VLMs, which are natural extensions of LLMs~\cite{team2024gemma, touvron2023llama, achiam2023gpt}, exacerbate existing issues with hallucinations by introducing an additional visual modality for potential errors~\cite{sun2023aligning, cui2023holistic, bai2024hallucination, bang-etal-2023-multitask}.
A significant line of work~\cite{li2023evaluating, jiang2024haleval, sun2023aligning, huang-etal-2024-lvlms, Guan_2024_CVPR, wu-etal-2024-autohallusion, wiles2024revisiting} relies on question-answering style verification, where an instruction-tuned VLM is expected to answer whether a particular fact about an image is valid. However, such an approach suffers from several limitations:
First, the lack of interpretability makes understanding where the model fails challenging. In the case of binary (yes/no) questions, there is a high probability (50\%) of the model answering correctly but for the wrong reason.
Second, the QA abilities of the models do not necessarily translate into captioning capabilities~\cite{Kaul2024THRONEAO}.

The alternative approach is to check if the facts mentioned in a model's output align with what is shown in the input images.
So far, this has been limited to the presence of objects defined in classic captioning datasets with ground-truth annotations. For instance, the seminal work, CHAIR~\cite{objectHallucination}, is strictly bound to the MS COCO dataset ~\cite{lin2014microsoft} as it requires a dataset with paired captions and object-label annotations. CHAIR extracts nouns from a predicted caption using traditional NLP techniques and verifies their presence in an image using ground-truth object annotations.
THRONE~\cite{Kaul2024THRONEAO}, ALOHa~\cite{petryk2024aloha} and VALOR-BENCH~\cite{qiu2024valor} improve on CHAIR by employing an LLM to parse generated free-form captions. While those methods can handle concepts outside of MS COCO classes, they can only do so to a limited extent. During parsing, THRONE only considers a closed set of objects specifically defined within a dataset (MS COCO or Object365~\cite{shao2019objects365}). VALOR-BENCH~\cite{qiu2024valor} only operates on carefully selected small subset of images from Visual Genome~\cite{krishna2017visual}.   
ALOHa relaxes CHAIR's string matching with a text similarity between annotated and predicted objects yet also assumes access to an exhaustive list of human-annotated references in the image.
In contrast, OpenCHAIR~\cite{ben-kish-etal-2024-mitigating} introduces its dataset to measure hallucination by generating images with a text-to-image model~\cite{podell2023sdxl} encompassing concepts different from MS COCO. However, the overall approach still relies on ground-truth references and since the accompanying captions are in MS COCO style, they are very short and typically describe a single object. 
In our work, we aim to measure the factuality of captions beyond predefined concepts, datasets and human annotations, by creating a \textit{reference-free}, open-vocabulary metric that also incorporates both precision and recall.
Crucially, our metric being reference-free allows for applicability to filtering of large-scale pretraining datasets, as discussed next.

\paragraph{Data filtering.}
Data filtering for contrastive image-text pretraining~\cite{radford2021learning} on web-scale data has shown it can lead to stronger models and improved training efficiency~\cite{fang2023data, gadre2024datacomp, xu2023demystifying, evans2024data, udandarao2024active}. Meanwhile,
data filtering for generative VLMs training has seen less attention, discussing data curation for instruction fine-tuning~\cite{wei2023instructiongpt, chen-etal-2024-vision, cao2023instruction}. 
Their primary focus, however, is to improve the instruction-following of the model and not on the factual quality of captioning outputs as we do here.
\citet{li-etal-2024-role} considers data curation for image captioning models by filtering or replacing examples with high training losses on small-scale datasets with human-annotated short captions. 
In contrast, our proposed metric can operate on large-scale pretraining datasets with noisy (VLM-generated) long captions in terms of their factuality; we show this leads to consistent improvements across a range of measures.
Crucially, this application is only possible as our metric is \textit{reference-free} and does not require ground-truth object annotations like prior works~\cite{petryk2024aloha, objectHallucination, Kaul2024THRONEAO}.

\section{Method}

We detail \ours{} in Sec.~\ref{sec:3_approach} to \textit{measure} open-vocabulary object factuality and then describe how our proposed metric can be used for data filtering to \textit{improve} a VLM's factuality in Sec.~\ref{sec3:data_filtering}.
Finally, Sec.~\ref{sec3:discussion} summarizes how \ours{} addresses the limitations of existing approaches.

\subsection{\ours~}
\label{sec:3_approach}
\paragraph{Overview.}
Our goal is to derive an interpretable, flexible, and reliable method for assessing the factuality of long captions. Because our approach should \textit{generalize to any caption}, regardless of its source dataset or vocabulary, we avoid reliance on dataset-specific terms.  We achieve this by verifying the correctness of visual entities and their attributes mentioned in a caption. Specifically, given a pair of image $x \,{\in}\, \mathbb{R}^{H \times W \times 3}$ and caption $y$, we first parse $y$ to extract a set of candidate entities. We then verify their presence in the image $x$ by running image grounding tools. To avoid promoting  non-descriptive captions we design a recall-based metric that compares candidate entities against reference entities extracted either from ground-truth captions (when reliable) or from grounding tools with a large vocabulary of concepts. The high-level overview of our approach is presented in Fig.~\ref{fig3:method}.

\paragraph{Caption parsing.} 
We start by parsing $y$ to extract candidate entities. To do so, we prompt an LLM to generate a list of objects mentioned in $y$ by specifically instructing an LLM to output all objects with their visual attributes, ignoring abstract concepts with no visual presence, such as \textit{sound} and \textit{atmosphere}, often present in long VLM descriptions. We provide the prompt details in the Appendix~\ref{app:llm_prompt}. We denote the resulting candidate entity set as $\mathcal{C}$. 

\paragraph{Entity grounding.}
Having obtained the set $\mathcal{C}$, we then validate the presence of each one of the candidate entities $c_i \in \mathcal{C}$ within image $y$. {Note that $c$ is a free-form text; thus, verifying the presence of associated concepts with no prior access to a vocabulary poses a challenge.
To address this, we first utilize a state-of-the-art open-vocabulary object detector~\cite{minderer2024scaling}.
Specifically, we feed each $c_i$ to a detection model as a separate query. We then define the candidate entity, $c_i$, as being grounded if its detection confidence score exceeds a threshold value. This yields an initial set of grounded entities, $\mathcal{G_D}$, where $\mathcal{G_D} \subset \mathcal{C}$.

In our empirical studies, we observed that long image captions often include much more details about surroundings and "stuff-like"~\cite{kirillov2019panoptic, forsyth1996finding} concepts (e.g., \textit{water}, \textit{sky}, \textit{concrete}) 
which object detectors typically miss. Thus, we incorporate an \textit{additional} grounding step using an open-vocabulary semantic segmentation model. Each candidate entity $c_i$ is input to a segmentation model to obtain a grounded set
$\mathcal{G_S} \subset \mathcal{C}$. Our final resulting set of all grounded entities is then $\mathcal{G} = \mathcal{G_D} \cup \mathcal{G_S}$.

\paragraph{OVFact Scoring.}
We leverage the parsing and grounding outputs to measure the overall factuality of caption $y$. Focusing on the specific challenges of long captioning, we assess two key aspects: precision and descriptiveness. While prior metrics \cite[e.g.,][]{objectHallucination} are more precision-focused, this is not as suitable for our setting, as this rewards short captions with fewer potential mistakes but few details. By unifying both aspects (in a reference-free manner), we can also apply our single metric for downstream applications (e.g. filtering).

\paragraph{Calculating Precision.}
We define the precision of a caption as the ratio between the count of the grounded entities $\mathcal{G}$ and candidate entities $\mathcal{C}$:

\begin{align}
\text{\ours{}}_\precisionsubs=\frac{|\mathcal{G}|}{|\mathcal{C}|}
\end{align}

\paragraph{Calculating Recall.}
To measure descriptiveness, we design an additional metric interpreted as traditional recall. We first obtain an approximate set of entities appearing in $y$. In an ideal scenario, i.e. when considering fully annotated datasets like MS COCO~\cite{lin2014microsoft}, one could use object-level annotations in the dataset as references $\mathcal{R}$. However, MS COCO is the only dataset to date with both human-annotated captions and object detection labels, although the captions are very short and thus not descriptive.

Assuming ground-truth captions are human-annotated and give an exhaustive description of a scene, a possible solution to this problem is extracting $\mathcal{R}$ from ground-truth captions by employing the parsing introduced before, which we propose in our approach.
However, if the ground-truth captions are unreliable, e.g. for VLM-generated datasets, we also consider a general case to obtaining $\mathcal{R}$ by prompting the grounding tools given a large enough vocabulary of concepts $\mathcal{V}$. 

We then measure the recall of reference entities $\mathcal{C}$ in $\mathcal{R}$.
To facilitate the open-vocabulary aspect of our approach (as entities in both sets are free-form texts), we first compute text embeddings for each $c_i \in \mathcal{C}$ and for all $r_j \in \mathcal{R}$; $f(c_i) \rightarrow \vec{c_i} \in \mathbb{R}^D$ and $f(r_j) \rightarrow \vec{r_j} \in \mathbb{R}^D$. We then calculate the similarity between feature representations from the two sets of entities with the cosine similarity $S_{ij} = \frac{\vec{r_j} \cdot \vec{c_i}}{|\vec{r_j}||\vec{c_i}|}$. 
We define recall by selecting a maximum similarity score for each reference entity and report as a final score an average of all $r_j \in \mathcal{R}$, that is:
\vspace{-0.5\baselineskip}
\begin{align}
\text{\ours{}}_{\avgsimmsubs} = \dfrac{1}{|\mathcal{R}|}\sum\limits_{j=1}^{|\mathcal{R}|}{\max\limits_{i=1}^{|\mathcal{C}|}{S_{ij}}} 
   \label{eq:avgsim}
\end{align}
\paragraph{Final metric.}
Finally, to encompass both the precision and descriptiveness of $y$ into one metric (our full \ours{}), we calculate our unified F1 score: 
\begin{align}
\text{\ours{}}_{F1} = \frac{ 2 \cdot \text{\ours{}}_\precisionsubs\cdot \text{\ours{}}_{\avgsimmsubs}}{\text{\ours{}}_\precisionsubs + \text{\ours{}}_{\avgsimmsubs}}
\end{align}

\subsection{\ours{} for Data Filtering}
\label{sec3:data_filtering}

The generalized case of \ours{} (Fig.~\ref{fig3:method}) can be completely reference-free.
This is particularly beneficial, as it can be applied to any set of image and caption pairs and is not limited to only clean, fully-labelled datasets.
This becomes particularly important when considering a growing number of LLM-generated caption datasets~\cite{chen2023sharegpt4v, arai2025covla}.
One important application of \ours{} is data filtering, where our metric serves as a scoring function for pruning incorrect samples, e.g. including hallucinated objects. 

Consider a dataset of (image $x$, caption $y$) pairs, where $y$ is generated at scale using a large VLM, and this data is intended to help train other models, as done in \cite{chen2023sharegpt4v}. Using the process described in previous sections, our data filtering approach consists of extracting \fours{} scores for each ($x$, $y$) pair. Because our method is \textit{reference-free}, we do not need additional human-generated ground-truth object labels. We then sort the pairs based on their \fours{} and select top X\% depending on the assumed data pruning ratio. This simple technique can result in a significantly improved performance in factuality for long captioning models without compromising descriptiveness. We demonstrate it experimentally in Sec.~\ref{sec4:data_filtering}.

\subsection{Discussion}
\label{sec3:discussion}

Having introduced \ours{}, we now revisit closely related work to highlight key differences.
\para{CHAIR}~\cite{objectHallucination} is a popular metric to measure hallucinations, but is strictly bound to the COCO dataset, both for the instance-level CHAIRi$=\frac{|\{\text{hallucinated objects}\}|}{|\{\text{all objects mentioned\}}|}$ and caption-level CHAIRs $= \frac{|\{\text{sentences with hallucinated object}\}|}{|\{\text{all sentences\}}|}$.
Moreover, CHAIR can be artificially improved by simply not predicting any objects. 
This occurs when a caption genuinely contains no objects or when predicted entities fall outside the COCO vocabulary.

\para{THRONE}~\cite{Kaul2024THRONEAO} improves CHAIR's string matching with LLM-based parsing, which can handle free-form captions.
However, the parsing \textit{output} is still limited to categories defined in the evaluation set (the LLM is prompted about a predefined list of objects).
While THRONE could be extended to datasets other than MS COCO, it would require annotated object detection labels for the final score, rendering it unsuitable for reference-free or object-annotation-free setups.
In contrast to CHAIR, THRONE assesses both precision and recall of a caption as we do in this work.

\begin{figure}
    \centering
    \includegraphics[width=\linewidth, trim={8em, 20.5em, 14.5em, 2.5em}, clip]{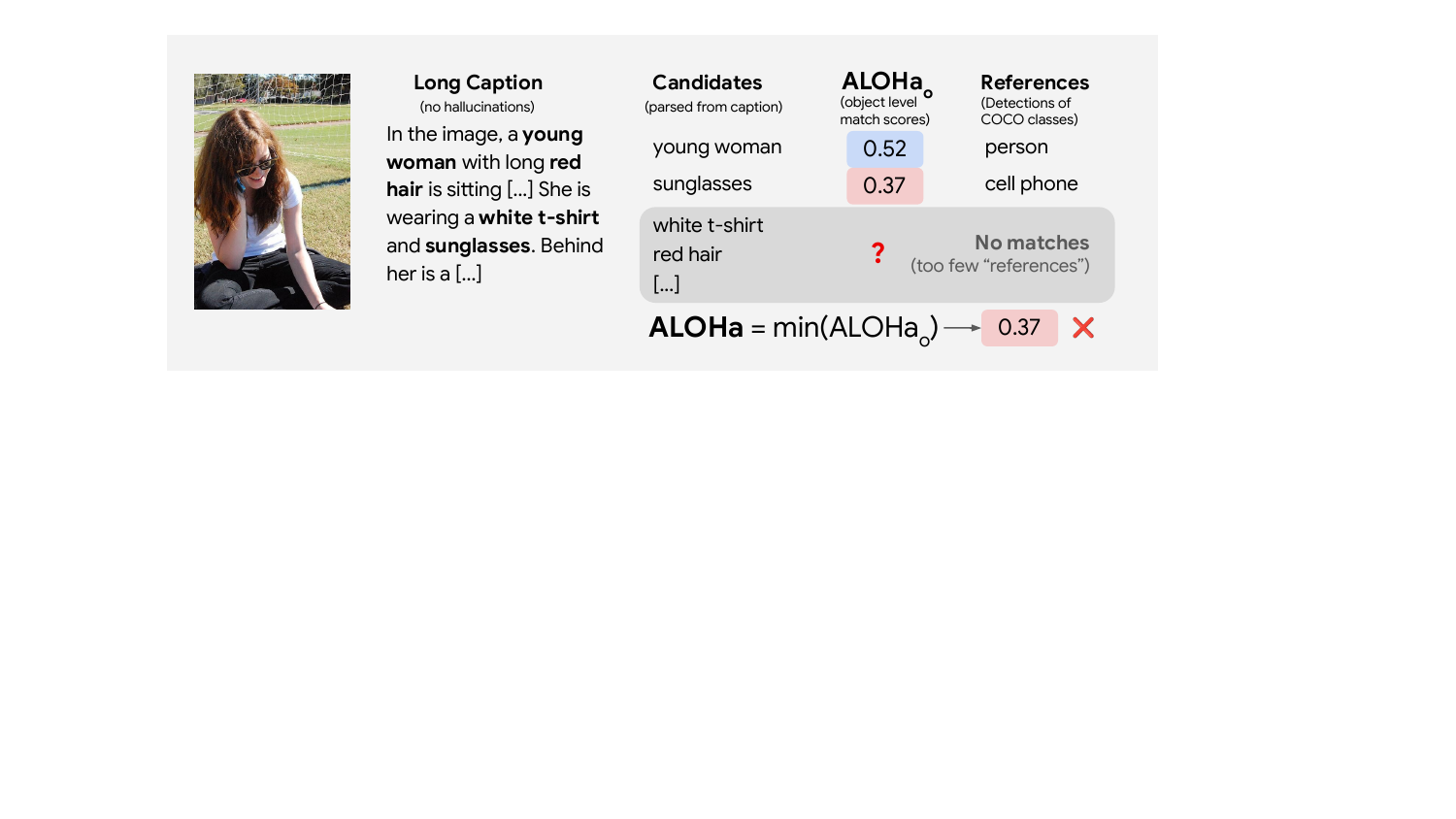}
    \vspace{-1.5\baselineskip}
    \caption{\textbf{Limitations of ALOHa in a reference-free setting.} We present an example from ShareGPT4v (i.e., with no access to ground-truth references). ALOHa's reference matching forces comparisons of detected classes with what has been described, an issue exacerbated in a reference-free setting when ground-truth human annotations of objects are unavailable. Our method is more robust to such cases.}
    \vspace{-\baselineskip}
    \label{fig4:aloha}
\end{figure}

\para{ALOHa}~\cite{petryk2024aloha} also addresses the closed-vocabulary limitation of CHAIR by using similarity scores between text embeddings of reference and candidate entities.
As in our approach, a language model is used to parse the candidate entities.
However, the reference set is constructed from ground-truth reference captions and additional object detections~\cite{carion2020end}. 
In contrast to our approach, ALOHa performs Hungarian matching~\cite{Kuhn1955Hungarian} on the similarity matrix, $S$, between candidates and references.
This difference is significant because Hungarian matching enforces a one-to-one correspondence between candidates and references and is therefore particularly sensitive to having an exhaustive and precise list of references as illustrated in Fig.~\ref{fig4:aloha}:
If there are not enough references to match with the candidates (i.e., $|\mathcal{R}| < |\mathcal{C}|$), ALOHa will ignore the unmatched candidates, leading to a misleading score.
Moreover, when $\mathcal{R}$ is incomplete, the one-to-one matching will force associations between unrelated concepts (in Fig.~\ref{fig4:aloha}, ``sunglasses'' is forced to match to reference ``cell phone'').
Finally, ALOHa favors shorter captions, as fewer candidates have a higher chance of being correctly matched to their references. 
These limitations of ALOHa are particularly evident when using it as a metric for data filtering, which we show experimentally in the next section.

\section{Experiments}
\vspace{-0.3\baselineskip}
\label{sec:4_exp}

In this section, we first describe our overall experimental setup and technical details in Sec.~\ref{sec4:exp_setup}. Next, in Sec.~\ref{sec4:verification}, we discuss the verification of \ours{} as method for \textit{measuring} factuality including ablation studies of its components. Finally, in Sec.~\ref{sec4:data_filtering}, we present results for \textit{improving} factuality, applying \ours{} for data selection. 

\subsection{Experimental setup}
\label{sec4:exp_setup}
\paragraph{Implementation details.}  We use the state-of-the-art open-source LLM Gemma2-27b~\cite{team2024gemma} for caption parsing and OWL-ViTv2~\cite{minderer2024scaling} and OpenSeg~\cite{openseg} for entity grounding. We consider an extensive vocabulary $\mathcal{V}$ of open-vocabulary concepts (as discissed in Sec.~\ref{sec:3_approach}). Specifically, we take the union of concepts from standard datasets, i.e. Visual Genome \cite{krishna2017visual}, LVIS \cite{gupta2019lvis}, Open Images~\cite{kuznetsova2020open} and Objects365~\cite{shao2019objects365} resulting in a total of 2792 unique concepts. To encode entities for \avgsim{} computation, we use the SigLIP-So400m/14~\cite{zhai2023sigmoid} text encoder. Additional details are in Appendix~\ref{app:experimental_details}.

\begin{table}[t]
\vspace{-\baselineskip}
\small
\centering
\begin{tabular}{l|cc}
\toprule
& DOCCI & Loc. Nar. \\

\midrule
\textbf{Gemma parsing}& 0.97 & 0.97 \\ 
CHAIR parsing & 0.02 &  0.28 \\
POS + CHAIR & 0.79 & 0.58 \\
\hline
\textbf{Entity grounding} & 0.72 & 0.79 \\
\quad w/o detection & 0.46 & 0.67 \\
\quad w/o segmentation & 0.62 & 0.64  \\
\bottomrule
\end{tabular}
\vspace{-0.5\baselineskip}
\caption{\textbf{Validation of OVFact components.} We measure the specificity of each of the steps in our method on 100 DOCCI \textit{qual dev} split and 100 random samples from Localized Narratives \textit{train} set. See Sec.~\ref{sec4:verification}.}
\vspace{-1.0\baselineskip}
\label{tab4:metric_val}
\end{table}

\paragraph{Datasets and Evaluation metrics.}
To validate the effectiveness of our approach, we conduct experiments on multiple captioning datasets with different levels of complexity.

\noindent \textbullet~\textbf{ShareGPT4v} ~\cite{chen2023sharegpt4v} is a large dataset of automatically generated long captions by GPT4v~\cite{achiam2023gpt}. It consists of approximately 100k samples of an average length of 950 characters.
We use the ShareGPT4v dataset as the primary \textit{training} dataset for data selection experiments as the VLM-generated captions are noisy, but widely adopted for training~\cite{liu2024improved, lu2024deepseek, tong2024cambrian}.

\noindent \textbullet~\textbf{DOCCI}~\cite{onoe2024docci} is a recent human-annotated dataset of detailed, high-quality captions (average length of 642 characters), with images sourced from voluntarily contributed, private archives (guaranteed to be no overlap with VLM training sets). Its caption length and diversity mean it is the most challenging published dataset for long, detailed captions. We select its  \textit{test} set with 5k samples as our downstream evaluation set.

\noindent \textbullet~\textbf{Localized Narratives}~\cite{PontTuset2022locnar} consists of long, human-annotated captions which average 206 characters.
We report downstream evaluation of our method 
on the COCO \textit{validation} split to complement our analysis further. 

\noindent \textbullet~\textbf{MS COCO}~\cite{lin2014microsoft} is the standard for measuring hallucinations as it includes bounding box annotations for 80 different object classes. However, captions are short (only 52 characters on average).
Following standard practice~\cite{objectHallucination, Kaul2024THRONEAO, petryk2024aloha}, we report CHAIR on the \textit{val} split to show generalization of our method to prior, standard settings.

\subsection{\ours{} for Measuring Factuality}
\label{sec4:verification}

\begin{table}[t]
\vspace{-\baselineskip}
\small
\centering
\begin{tabular}{l|
>{\centering\arraybackslash}m{2.7em}
>{\centering\arraybackslash}m{4.6em}|
>{\centering\arraybackslash}m{4.0em}
}

\toprule
& ALOHa & \precision{} & \avgsim{} \\
\midrule

Human agreement & 48.6\% & 72.1\%  &  80.7\% \\

\bottomrule
\end{tabular}
\vspace{-0.5\baselineskip}
\caption{\textbf{\ours{} improves alignment with human judgments.} 
We extend the analysis by \citet{onoe2024docci} on DOCCI over 4 VLMs, observing that \precision{} shows higher agreement with human judgment compared to prior work (ALOHa).
We highlight that (1) \ours{} is \textit{reference-free}, while ALOHa here is given references from ground-truth captions, and (2) our method also captures recall (descriptiveness) and shows high agreement with human judgments (\avgsim{}). See Appendix~\ref{app:human_alignment} for further details.
}
\vspace{-1\baselineskip}
\label{tab4:metric_val_human_study}
\end{table}

\begin{figure*}[!t]
    \vspace{-1.5\baselineskip}
    \centering
        \includegraphics[width=\linewidth]{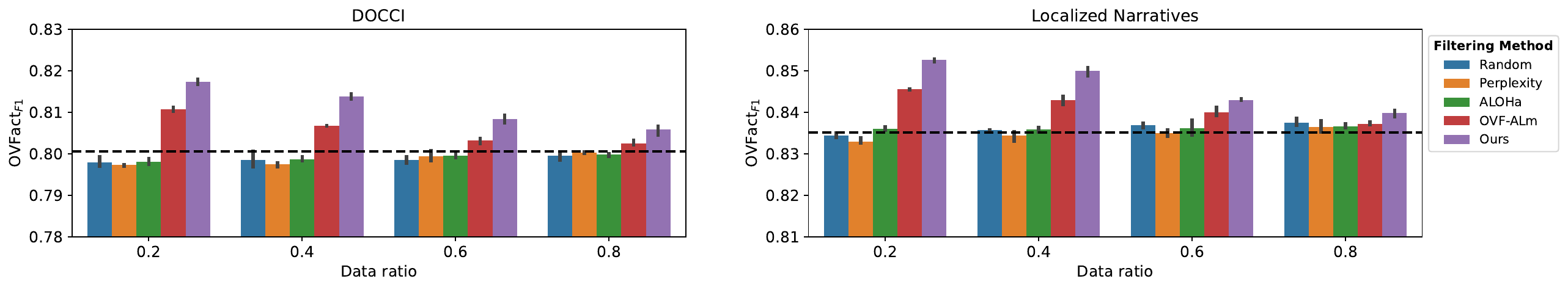}
    \vspace{-2\baselineskip}
    \caption{\textbf{Data filtering comparison measured with \fours}. We present the results of ShareGPT4v filtering for different data ratios, with \textit{zero-shot evaluations} on DOCCI \textit{test} and Localized Narratives \textit{val} sets averaged over 3 training runs. '\textcolor{black}{-\hspace{0.1mm}-\hspace{0.1mm}-}' indicates model trained on \textit{full} ShareGPT4v dataset. \ours{} significantly improves ($\uparrow$) on factuality over prior work, particularly as we filter more examples (resulting in a smaller, higher-quality training set). We also report an OVF-ALm baseline, which shows the impact of using ALOHa's (brittle) matching with our OVFact method. We highlight our gains in factuality come without compromising descriptiveness or quality.
    }
    \label{fig4:main_results}
\end{figure*}


\pgfplotstableread{Fig/data/chairi.dat}\tablea 

 \pgfplotstableread{Fig/data/chairi_rand.dat}\tableaa

\pgfplotstableread{Fig/data/chairs.dat}\tableb
\pgfplotstableread{Fig/data/chairs_rand.dat}\tableab

\pgfplotstableread{Fig/data/recall.dat}\tablec

\pgfplotstableread{Fig/data/recall_rand.dat}\tableac

\pgfplotsset{every axis legend/.append style={font=\tiny}}



\definecolor{Ours}{RGB}{147,115,178} 
\definecolor{Aloha_ext}{RGB}{192,61,62}
\definecolor{Aloha_van}{RGB}{58,146,59}
\definecolor{Perplexity}{RGB}{225,129,43}
\definecolor{Random}{RGB}{50,116,161}

\begin{figure*}[t!]
\centering
\begin{tikzpicture}
\begin{axis}[
    title={},
    xlabel={\footnotesize Data ratio},
    ylabel={ \footnotesize CHAIRi $\leftarrow$},
    ylabel style={yshift=-10pt},
    legend to name=sharedLegend,
    width=0.31\textwidth,
    anchor=east,
    name=plot1,
    xmin=0.14,
    xmax=0.93,
    ymin=6.8,
    tick label style={font=\footnotesize},
    label style={font=\footnotesize}
]
\addplot[mark=*,color=Ours] table[x=ds_ratio,y=OVFact_v2] {\tablea};
\addplot[mark=*,color=Perplexity] table[x=ds_ratio,y=pplx_low] {\tablea};
\addplot[mark=*,color=Aloha_van] table[x=ds_ratio,y=aloha_vanila] {\tablea};
\addplot[mark=*,color=Aloha_ext] table[x=ds_ratio,y=aloha_ext] {\tablea};
\addplot[mark=*,color=Random] table[x=ds_ratio,y=random] {\tablea};
\addplot[color=Random, dashed, fill=Random!20, opacity=0.5] table[x=ds_ratio,y=random] {\tableaa} \closedcycle;
\legend{OVFact\_v2, pplx\_low, aloha\_vanila, aloha\_ext, random}
\end{axis}
\label{fig4:chairi}
\end{tikzpicture}
\begin{tikzpicture}
\begin{axis}[
    title={},
    xlabel={ \footnotesize Data ratio},
    ylabel={ \footnotesize CHAIRs $\leftarrow$},
    ylabel style={yshift=-11pt},
    legend to name=sharedLegend,
    width=0.31\textwidth,
    at={(0.3\textwidth,0)},
    xmax=0.93,
    xmin=0.14,
    tick label style={font=\footnotesize},
    label style={font=\footnotesize}
]
\addplot[mark=*,color=Ours] table[x=ds_ratio,y=OVFact_v2] {\tableb};
\addplot[mark=*,color=Perplexity] table[x=ds_ratio,y=pplx_low] {\tableb};
\addplot[mark=*,color=Aloha_van] table[x=ds_ratio,y=aloha_vanila] {\tableb};
\addplot[mark=*,color=Aloha_ext] table[x=ds_ratio,y=aloha_ext] {\tableb};
\addplot[color=Random, dashed, fill=Random!20, opacity=0.5] table[x=ds_ratio,y=random] {\tableab} \closedcycle;
\addplot[mark=*,color=Random] table[x=ds_ratio,y=random] {\tableb};
\legend{OVFact\_v2, pplx\_low, aloha\_vanila, aloha\_ext, random}
\end{axis}
\label{fig4:chairs}
\end{tikzpicture}
\begin{tikzpicture}
\begin{axis}[
    title={},
    xlabel={ \footnotesize Data ratio},
    ylabel={ \footnotesize Recall $\rightarrow$},
    at={(0.66\textwidth,0)},
    ylabel style={yshift=-6pt},
    width=0.31\linewidth,
    ymin=59.8,
    ymax=61.8,
    xmax=0.93,
    xmin=0.14,
    tick label style={font=\footnotesize},
    label style={font=\footnotesize},
    legend style={
                at={(1.3,0.9)},  
                anchor=north,
                legend columns=1,  
                column sep=0em,    
                cells={anchor=west},
                draw=none,         
            },
]
\addplot[mark=*,color=Random] table[x=ds_ratio,y=random] {\tablec};
\addplot[mark=*,color=Perplexity] table[x=ds_ratio,y=pplx_low] {\tablec};
\addplot[mark=*,color=Aloha_van] table[x=ds_ratio,y=aloha_vanila] {\tablec};
\addplot[mark=*,color=Aloha_ext] table[x=ds_ratio,y=aloha_ext] {\tablec};
\addplot[mark=*,color=Ours] table[x=ds_ratio,y=OVFact_v2] {\tablec};
\addplot[color=Random, dashed, fill=Random!20, opacity=0.5] table[x=ds_ratio,y=random] {\tableac} \closedcycle;
\addlegendentry{Random}
\addlegendentry{Perplexity}
\addlegendentry{ALOHa}
\addlegendentry{OVF-ALm}
\addlegendentry{Ours}
\end{axis}
\label{fig4:rec}
\end{tikzpicture}

\vspace{-0.5\baselineskip}
\caption{\textbf{Generalization to CHAIR benchmark.} We also measure zero-shot evaluation on the CHAIR benchmark, highlighting how our filtering improves on traditional metrics and settings too. 
Light-blue shading marks variance of the \textit{random} baseline. Our filtering method results in fewer hallucinations at the instance and sentence level (CHAIRi, CHAIRs; \textbf{lower} $\downarrow$ is better), maintaining or improving recall (\textbf{higher} $\uparrow$ is better) of generated captions.
}
\label{fig4:chair}
\vspace{-\baselineskip}
\end{figure*}
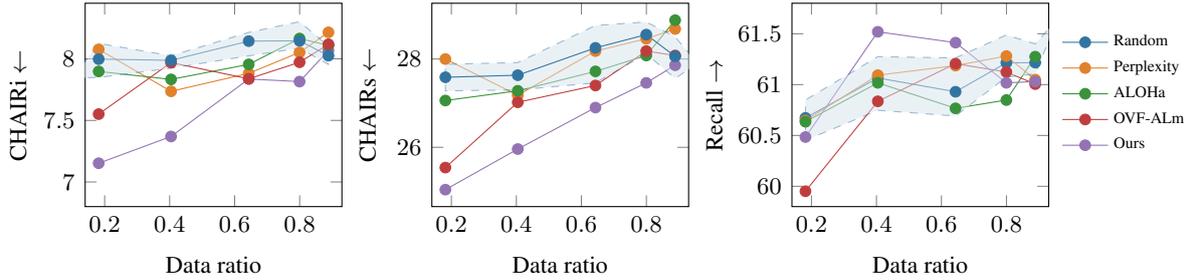

To assess the reliability of our proposed method, we validate its performance on manually annotated subsets from the DOCCI \textit{qual dev} split and 100 randomly sampled image-caption pairs from the Localized Narratives \textit{train} split. We manually annotate entities in ground-truth captions and then compute and measure the performance 
of each step in our approach, which we report in Tab.~\ref{tab4:metric_val}. First, we measure the specificity of the caption parsing stage, which is measured as the ratio of entities extracted by Gemma to annotated entities. We also show an ablation of Gemma parsing by comparing it to string matching on the limited CHAIR vocabulary (CHAIR parsing), as well as to a variant of string matching enhanced with POS Tagging ~\cite{honnibal2020spacy}. Our caption parsing demonstrates robust performance, achieving a specificity of 97\% on both datasets.

We further analyze the specificity of the entity grounding component discussed in Sec.~\ref{sec:3_approach}, calculated as the ratio of correctly grounded entities to all annotated entities in the ground-truth captions. Overall, we observe a difference in the specificity of entity grounding between the two considered datasets, most likely because the DOCCI descriptions are much more detailed, with numerous examples of specific objects. We also analyze in Tab.~\ref{tab4:metric_val} separate components of our grounding stage and conclude that the use of object detection and segmentation proves crucial for both datasets, producing a specificity improvement of up to 0.26 in DOCCI compared to the use of segmentation alone.

Finally, in Tab.~\ref{tab4:metric_val_human_study}, we extend the human study analysis from \citet{onoe2024docci} over four state-of-the-art long caption models applied to the DOCCI dataset and observe that our model aligns well with human preferences for both \precision{} and \avgsim{}. We also compare the results of ALOHa in the original setup with ground-truth references and observe that \precision{} is more aligned with human judgments compared to ALOHa.
We provide details and results of this analysis in Appendix~\ref{app:human_alignment}.

\subsection{\ours{} for Improving Factuality}
\label{sec4:data_filtering}

\paragraph{Experimental setup.}
For our data filtering experiments, we consider how \ours{} can be used to score and filter a large-scale, noisy training dataset with generated VLM captions (ShareGPTv, Sec.~\ref{sec4:exp_setup}), selecting the top $X$\% with the highest \ours{}. We validate our approach by considering how supervised fine-tuning on these filtered sets can improve a representative open-source state-of-the-art VLM, PaliGemma 2 ~\cite{steiner2024paligemma}, on downstream zero-shot long caption evaluations (details Sec.~\ref{sec4:exp_setup}).
We report here the results with PaliGemma 2 (3B) model, and discuss results with additional sizes in Appendix~\ref{app:model_study} and give more training details in Appendix~\ref{app:experimental_details}.

\paragraph{Filtering Baselines.}
To our knowledge, \ours{} is the first open-vocabulary, reference-free method for measuring factuality in long captions, and prior methods that require ground-truth human references inherently cannot work in this data filtering setting, where only the noisy VLM-generated captions are available. Nonetheless, in addition to \textbf{Random} filtering, we consider the following:

\noindent \textbullet~\textbf{Perplexity}: we obtain a perplexity score predicted by the base, pre-trained PaliGemma 2 model for each sample. We then select samples with the lowest perplexity score and discard high-loss samples similarly to~\citet{li-etal-2024-role}.

\noindent \textbullet~\textbf{ALOHa}: we implement a reference-free variation of ALOHa, where we only rely on a set of detected objects for each image. For consistent comparison, we apply the same base models/tools as in \ours{} (Gemma 2, etc.) with vocabulary, matching method, etc. settings from the original paper. Samples with the highest ALOHa ``caption-level'' score \cite{petryk2024aloha} are selected for training.

\noindent \textbullet~\textbf{\ours{} + ALOHa matching (OVF-ALm)}: to better measure the limitations of ALOHa's matching in our reference-free setting (see Sec.~\ref{sec3:discussion}), we introduce stronger hybrid baseline OVF-ALm, where initial parsing and grounding is with our \ours{} full open-vocabulary concept set $\mathcal{V} \rightarrow \mathcal{R}$, and we apply ALOHa's matching algorithm at the end to obtain the final caption-level score for selection.

\paragraph{Results.}
Fig.~\ref{fig4:main_results} presents the results of applying various data filtering methods and their downstream performance on DOCCI and Localized Narratives datasets, evaluated using our proposed metric (\fours{}). Results are averaged over three independent training runs for each method. The dashed line indicates the performance of a model trained on the full ShareGPT4v dataset. Our filtering method consistently outperforms all other methods across all data ratios on both datasets, with the most pronounced improvements observed at lower data ratios, suggesting that the original dataset contains a large amount of noise.
\begin{figure}[t]
    \centering
    \includegraphics[width=\linewidth]{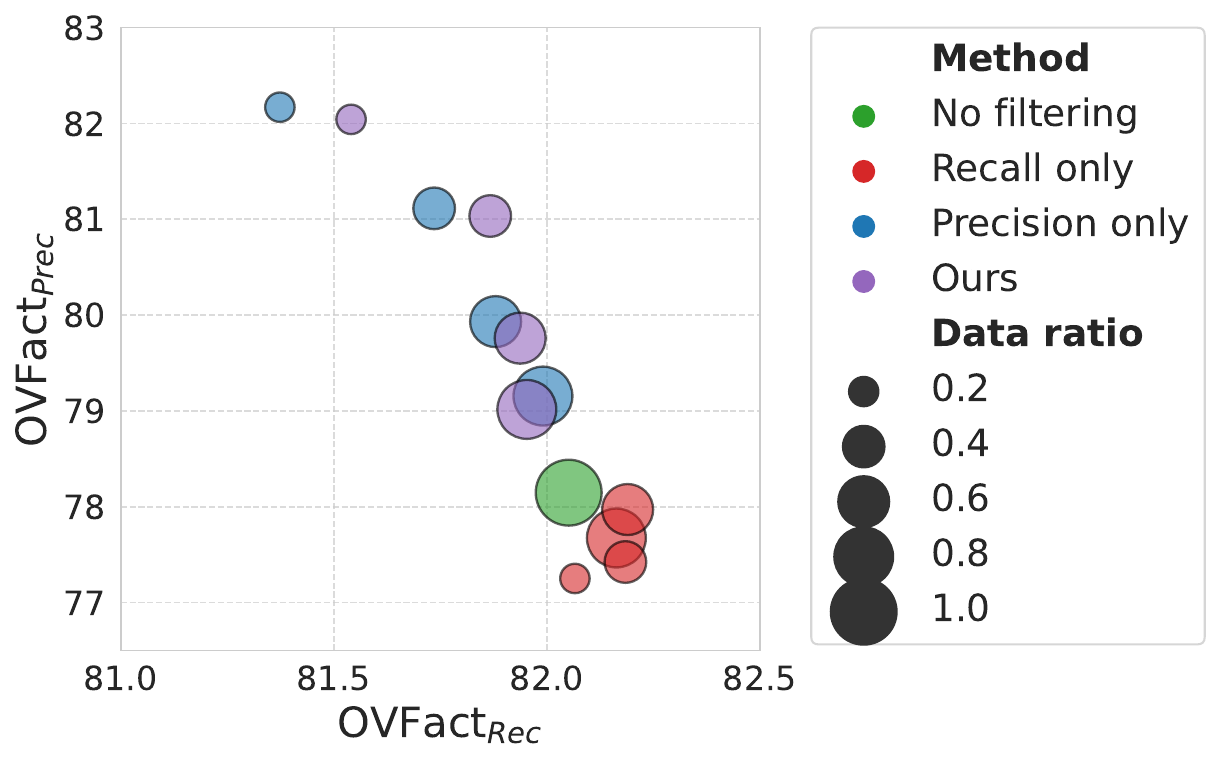}
    \vspace{-1.5\baselineskip}
    \caption{\textbf{Data-filtering ablation study}. We train PaliGemma 2 on filtered subsets of ShareGPT4V (with ablations of our method), and evaluate models zero-shot on the DOCCI test set. Note, top-right corner is the best.
    Filtering with \textit{full} \ours{} (purple) yields models with best trade-off between precise and descriptive captions.
    }
    \vspace{-\baselineskip}
    \label{fig4:ablation}
\end{figure}
Interestingly, both perplexity- and ALOHa-filtering baselines perform comparably to random sampling. ALOHa's strict reliance on reference data, makes it unsuitable for data filtering. Our strong hybrid baseline OVF-ALm gives a positive signal for filtering yet still underperforms our proposed approach due to inherent limitations (Sec.~\ref{sec3:discussion}).

We also analyze the impact of our data selection method on the traditional CHAIR setting, to assess generalization. Fig. \ref{fig4:chair} presents the performance of models trained on varying sizes of ShareGPT4v data, evaluated on both (sentence) CHAIRs and (instance) CHAIRi in a zero-shot manner. In addition, we report recall following standard protocols in prior work \cite{favero2024multi}. Our method achieves the best results on CHAIR, particularly when using smaller training datasets. 
Notably, improved hallucination scores are accompanied by comparable or even higher recall values, demonstrating the effectiveness of our method in balancing the precision and recall of generated captions.

We further analyze the impact of our data selection method on the performance of VLMs with varying parameter sizes and observe that our data selection consistently improves performance across all model sizes, as detailed in Appendix~\ref{app:model_study}. 

\paragraph{Ablation studies.}
We study other variants of our filtering method in Fig.~\ref{fig4:ablation}, where we present a trade-off between \precision{} and \avgsim{} measured with \ours{} when evaluating zero-shot on DOCCI test set. In particular, we experiment with \textit{Precision only} approach, where we run our data selection process based on \precision{} of original captions in ShareGPT4v. Similarly, we apply the same process but with \avgsim{}, which we denote \textit{Recall only}. We compare all the variants against the full ShareGPT4v dataset (\textit{No filtering}). First, we notice that \textit{Recall only} approach results in slight improvement on \avgsim{} over full data training, yet clearly outperforms other variants in terms of \precision{}. On the other hand, \textit{Precision only} filtering decreases \avgsim{} with the decrease in training data size. Finally, filtering with 
\fours (full \ours{}) achieves a sweet spot between the two aspects, with a model trained with 40\% data improving \precision{} by 3 points with respect to the model trained with the entire dataset, yet staying almost on par in terms of \avgsim{}.

\paragraph{Qualitative examples.} We include qualitative analysis of our method in the Appendix~\ref{app:qualitative_examples}, including comparative analysis of both \textit{filtered examples} from ShareGPT4V and of the outputs of the \textit{models} trained on our higher-quality filtered set.

\paragraph{Human evaluation.}
Since traditional caption metrics (BLEU, CIDEr, etc.) are not well-suited for assessing general quality for long captions, as noted by \citep{onoe2024docci}, we also perform a side-by-side comparison of captions generated by a model trained on our higher-quality filtered set (the top 20\%, ranked by our method) compared with one trained on the \textit{full} dataset. Our results indicate that humans preferred the outputs from the model trained on the filtered subset \textbf{68.2\%} of times, even with 5x less training data (see Appendix~\ref{app:traditional_metrics}).

\section{Conclusions}
We introduced \ours{}, a novel method for \textit{measuring} caption factuality that leverages open-vocabulary visual grounding and tool-based verification. Unlike previous metrics, OVFact is not dataset-specific and effectively evaluates the factuality of long, descriptive captions.
We further demonstrated how OVFact can be used to \textit{improve} the factuality of VLMs by using it as a metric for data filtering.
We show that models trained on an OVFact-filtered subset (2.5-5x smaller) of a large-scale, noisy (VLM-generated) pretraining set meaningfully improved factuality precision without sacrificing caption descriptiveness when evaluating zero-shot on various downstream datasets.

\section*{Acknowledgements}

We would like to thank Yasumasa Onoe for insightful discussions and sharing the results of human evaluations of DOCCI. We would also like to thank Emanuele Bugliarello for his feedback on the manuscript.
\section{Limitations \& Future Work}

\para{Potential risks and considerations for broader impacts.}
Large Vision-Language Models present several significant challenges and risks that require careful consideration \cite{bommasani2021opportunities,mitchell2019model,gebru2021datasheets}. These models can potentially propagate harmful biases present in training data \cite{howard2024socialcounterfactuals}. From a technical perspective, VLMs may hallucinate details or generate false visual interpretations, which could be problematic in high-stakes, safety-critical applications like medical imaging or autonomous systems \cite{chen2024detecting,objectHallucination}. Our work is a step towards both measuring and improving the factuality of VLMs. We believe that an automated way to detect and improve hallucinations is an important challenge with a growing amount of AI-generated content and datasets, and that a version of our method can be helpful for settings where AI is deployed at scale and it would be necessary to assess factuality of VLM outputs in real-time as part of a larger system. However, our work, as it is presented here, should be considered an academic exploration into this direction, and extensions for deployment settings will require additional work to ensure proper mitigation of potential risks are in place. We detail additional aspects for consideration for both the ``measurement'' and ``improvement'' aspects of our work:

\para{Measuring Factuality.} Our method is motivated by the limitations of prior work, and succeeds in improving several aspects (e.g., reducing the dependence on ground-truth references noted in \citet{petryk2024aloha}, etc.). However, our \ours{} method operates by leveraging base vision and language models as tools, and while we worked to reduce the impact of potential errors in individual tools by composing them together (e.g., combining both detection and segmentation models to leverage their relative strengths), our method inherits and remains sensitive to their intrinsic limitations. We characterize the quantitative performance of these components in Sec.~\ref{sec4:verification}, but to expand, we observe that challenging visual inputs (e.g., with heavily occluded, very fine-grained, or distant objects or parts) would often prove difficult for tools to properly ground. We anticipate that future models for these tasks, with further refinements in pre-training data and architecture designs, will help to alleviate this, as these are an active area of research \cite{myers2024spin}. Similarly, the pre-training domain for these tools is important: while our datasets were on general images, to apply our general framework to specialized domains (e.g., medical image analysis \cite{chen2024detecting}) more bespoke models will be necessary. Further, the pre-training data for these models will be potential sources of bias inherited by the general framework, and will need to be considered for deployment settings. Finally, we also consider here the limitations in the definition and scope of ``factuality'' we consider here: as noted in the footnote on the first page, our focus was on \textit{object-level} caption factuality, whether noun phrases in VLM generated text were both accurate to (precision) and descriptive of (recall) the original visual input. We believe expanding this scope to include additional attributes (e.g., described inter-object relationships and dependencies) and domains (e.g., verbs, motion, adverbial phrases as particularly important in videos) are exciting directions for future work. Similarly, investigating factuality in the context of external databases of multimodal visual knowledge \cite[e.g.,][]{mensink2023encyclopedic} would be an interesting direction for future work.

\para{Improving Factuality.} We investigate one potential direction to improve factuality: by leveraging our new reference-free, open-vocabulary method, we are able to apply our method to filtering large scale pretraining data for a set widely adopted by the community \cite{liu2024improved, lu2024deepseek, tong2024cambrian}, showing our models achieve substantial gains on factuality and quality, with 2.5 - 5x less training data. However, while this filtered data is higher-quality, our method would need to be integrated as part of a larger framework to facilitate further corrections (e.g., by human annotators) to edit and further refine the captions themselves -- this would an interesting direction for future work. Filtering data has been recently shown in other contexts (contrastive vision-language models, e.g., for CLIP-style models for image retrieval) to unintentionally reduce accuracy in low-resource domains, e.g., in multilingual and multicultural contexts not well-represented by existing (largely English-centric) multimodal evaluations \cite{pouget2024nofilter}. Exploring how our method could potentially complement this analysis in long captioning settings would be an interesting direction for further study.
Finally, there is a growing body of literature on measuring factuality of generative visual models~\cite[e.g.,][]{wiles2025one, lee2024holistic}; while our focus is on long caption generation, these models generate visual data (images, videos) with similar base architectural designs. We believe that exploring connections of our method and related work in that space could prove to be a fruitful direction.

\bibliography{custom}

\clearpage
\appendix
\section{ Appendix}
\label{sec:appendix}

\setcounter{table}{0}
\renewcommand{\thetable}{A\arabic{table}}
\setcounter{figure}{0}
\renewcommand{\thefigure}{A\arabic{figure}}

\subsection{\ours{} and Human Alignment}
\label{app:human_alignment}

We include here the details and the results of the human alignment analysis. We extend the analysis of human studies in DOCCI~\cite{onoe2024docci}. We compare 4 models: InstructBLIP~\cite{instructblip}, LLaVA-1.5~\cite{liu2023visual}, GPT4v~\cite{achiam2023gpt} and PaLI-5B~\cite{chen2023pali} fine-tuned on DOCCI.
We conduct side-by-side (S$\times$S) comparisons between all 
6 pairs of models (the original study focused on 3 of these), where participants are asked to provide a preference between model A or model B in two areas: Precision and Descriptiveness (Recall). We randomly swap models while displaying them to ensure that there is no side bias. 
Annotations are made by four different annotators, and each sample has at least two judgments. Participants select one of the three options: \textit{A is better}, \textit{Neutral}, \textit{B is better}.
We then compare for each sample whether human judgments for Recall correspond to \avgsim{} (\avgsim{} (model A) > \avgsim{} (model B) ), and the same for \precision{}, whether they correspond to human preference on the Precision scale. In other words, if we present outputs from two models $A$ and $B$ (randomly shuffled for each $A \times B$ pair), does our metric (the difference between OVFact($A$) and OVFact($B$)) match with what humans prefer, for precision and recall? 

Additionally, we compare with other baselines, particularly ALOHa, against Human preference in Precision, though notably we provide ALOHa access to references parsed from human-annotated ground-truth captions (not available to our reference-free metric) to make it a strong comparison point. We also consider the CLIP-Image-Score proposed in \cite{ge2024visual}, which measures factuality with an indirect approach: an image generation model is applied to the output caption under consideration, and CLIP similarity is measured between the original image and this generated one -- we implement this metric with the same image generation model (Stable Diffusion XL) as in the original paper.
Tab.~\ref{tab:app_human_st} details the results which complement Tab.~\ref{tab4:metric_val_human_study} from the main paper. We present S$\times$S comparisons between all combinations of pairs of discussed VLMs, and observe consistent improvements throughout the range of output styles and capabilities, indicating the general applicability of our method. We highlight that our method, with its grounding and explicit precision/recall measures, matches substantially better with human judgment than both ALOHa and CLIP-Image-Score\footnote{Image generation models have known issues with \textit{faithful} generation for long input prompts \cite{wiles2025one} and these results suggest this kind of indirect, multi-stage generation approach may not (yet) be well-suited for the large-scale filtering applications explored here.}.

\begin{table*}[t]
\small
\centering
\begin{tabular}{c|ccc|cc}
\toprule
 & \multicolumn{5}{c}{Agreement rate \% with Human Judgments} \\
 & \multicolumn{3}{c|}{\textit{Precision}}  &  \multicolumn{2}{c}{\textit{Recall}} \\
\midrule
S$\times$S model comparison  & ALOHa & CLIP-Image-Score & \precision{} & CLIP-Image-Score & \avgsim{} \\
\midrule
GPT4v $\times$ LLaVA-1.5 & 55.6 & 66.7 & 88.9 & 75.0 & 91.7 \\
GPT4v $\times$ PaLI-5B & 46.2 & 38.5 & 53.8 & 60.0 & 73.3 \\
GPT4v $\times$ InstructBLIP & 25.0 & 58.3 & 50.0 & 53.3 & 80.0 \\
LLaVA-1.5 $\times$ PaLI-5B & 58.8 & 58.8 & 82.3 & 70.0 & 90.0\\
LLaVA-1.5 $\times$ InstructBLIP & 50.0 & 40.0 & 70.0 & 25.0 & 75.0 \\
InstructBLIP $\times$ PaLI-5B & 54.5 & 54.5 & 90.9 & 30.0 & 69.2 \\
\midrule
\textbf{Overall} (Tab.~\ref{tab4:metric_val_human_study}) & 48.6 & 52.7 & 72.1 & 55.4 & 80.7   \\
\bottomrule
\end{tabular}
\caption{
\textbf{OVFact improves agreement with side-by-side (S$\times$S) human evaluations.} Here, we show the extended analysis from Tab.~\ref{tab4:metric_val_human_study}. We extend the initial human study from \citet{onoe2024docci}, which compared the outputs of a fine-tuned PaLI-5B~\cite{chen2023pali} against each of 3 other state-of-the-art models (GPT4v~\cite{achiam2023gpt}, InstructBLIP~\cite{instructblip}, and LLaVa-1.5~\cite{liu2024improved}).
We report the agreement rate (\%) of our method vs. the judgment of human annotators for which model's output has higher precision or recall (see Sec.~\ref{app:human_alignment}), and show results for all 6 pairs of the 4 models.
Our model significantly improves on capturing precision compared to the SOTA metric in prior work (ALOHa \cite{petryk2024aloha}), even when provided references from ground truth (GT) captions. In contrast, our metric is fully \textit{reference-free}. We also report the agreement between our method and human recall judgments, demonstrating a high level of agreement. 
We also include the results for  CLIP-Image-Score~\cite{ge2024visual}, which incorporates image generation models to indirectly measure hallucination - we find our method substantially improves over this technique as well.}

\label{tab:app_human_st}
\end{table*}

\subsection{Model size study}
\label{app:model_study}
\begin{figure}
    \centering
    \includegraphics[width=\linewidth]{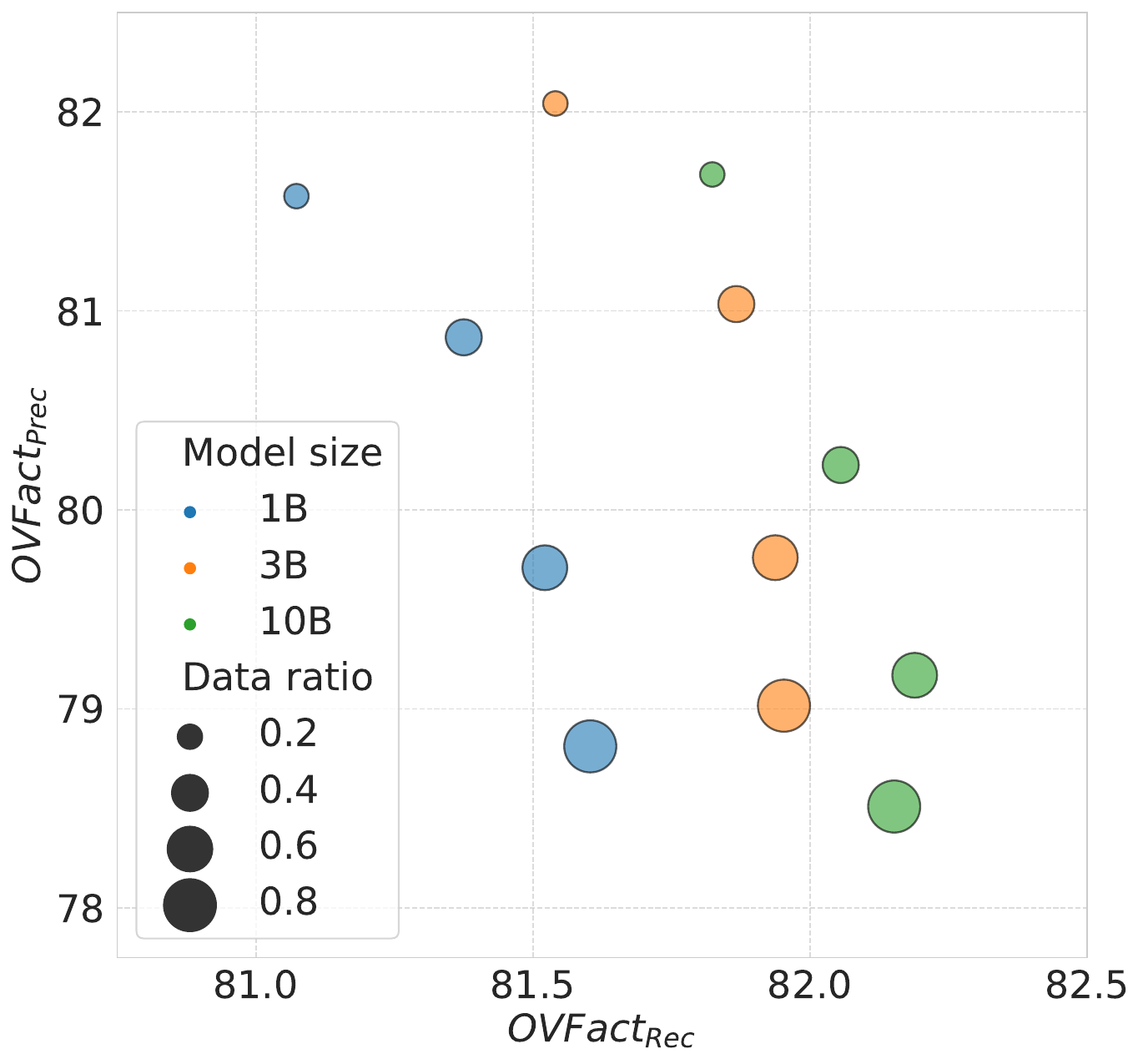}
    \caption{\textbf{Model size study} as a Precision-Recall curve. We analyze the impact of applying our data filtering strategy to various PaliGemma2 model sizes on DOCCI \textit{test} measured with \ours{}. Note that the best-performing models are closest to the top-right corner. Our method consistently improves models' performance across different model sizes.}
    \label{fig:app_model_study}
\end{figure}

In the main paper, our analysis was primarily on the 3B version of PaliGemma 2. We analyze the impact of our data selection method on the performance of Vision-Language Models (VLMs) with varying parameter sizes. Figure \ref{fig:app_model_study} presents the trade-off between \precision{} and \avgsim{} measured on the DOCCI test set for several PaliGemma2 variants. Across all model sizes, our data selection method improves performance. As the amount of training data increases, we observe a consistent trend: precision decreases while average similarity (\avgsim{}) increases. This trade-off is most pronounced for the smallest model, suggesting the importance of balancing precision and recall, especially for models with limited capacity.  Interestingly, the largest (10B parameter) model consistently achieves slightly higher \avgsim{} scores than smaller models, but often at the cost of slightly lower \precision.

\subsection{Qualitative examples}
\label{app:qualitative_examples}

\begin{figure*}[h!]
    \centering
    \includegraphics[width=0.56\linewidth, trim={3em, 14em, 23em, 1.5em}, clip]{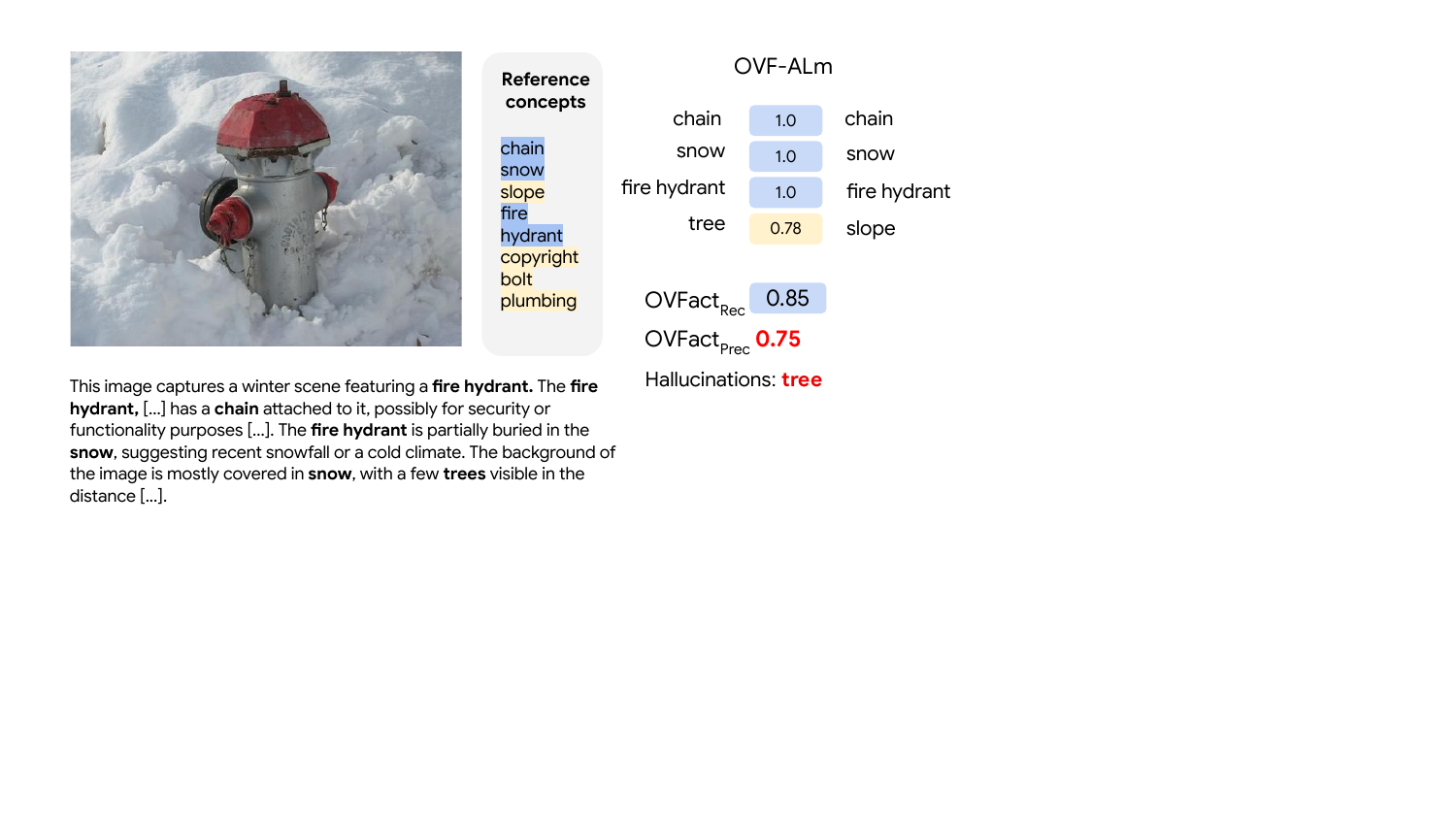}
    \includegraphics[width=0.43\linewidth, trim={3em, 8em, 26em, 0em}, clip]{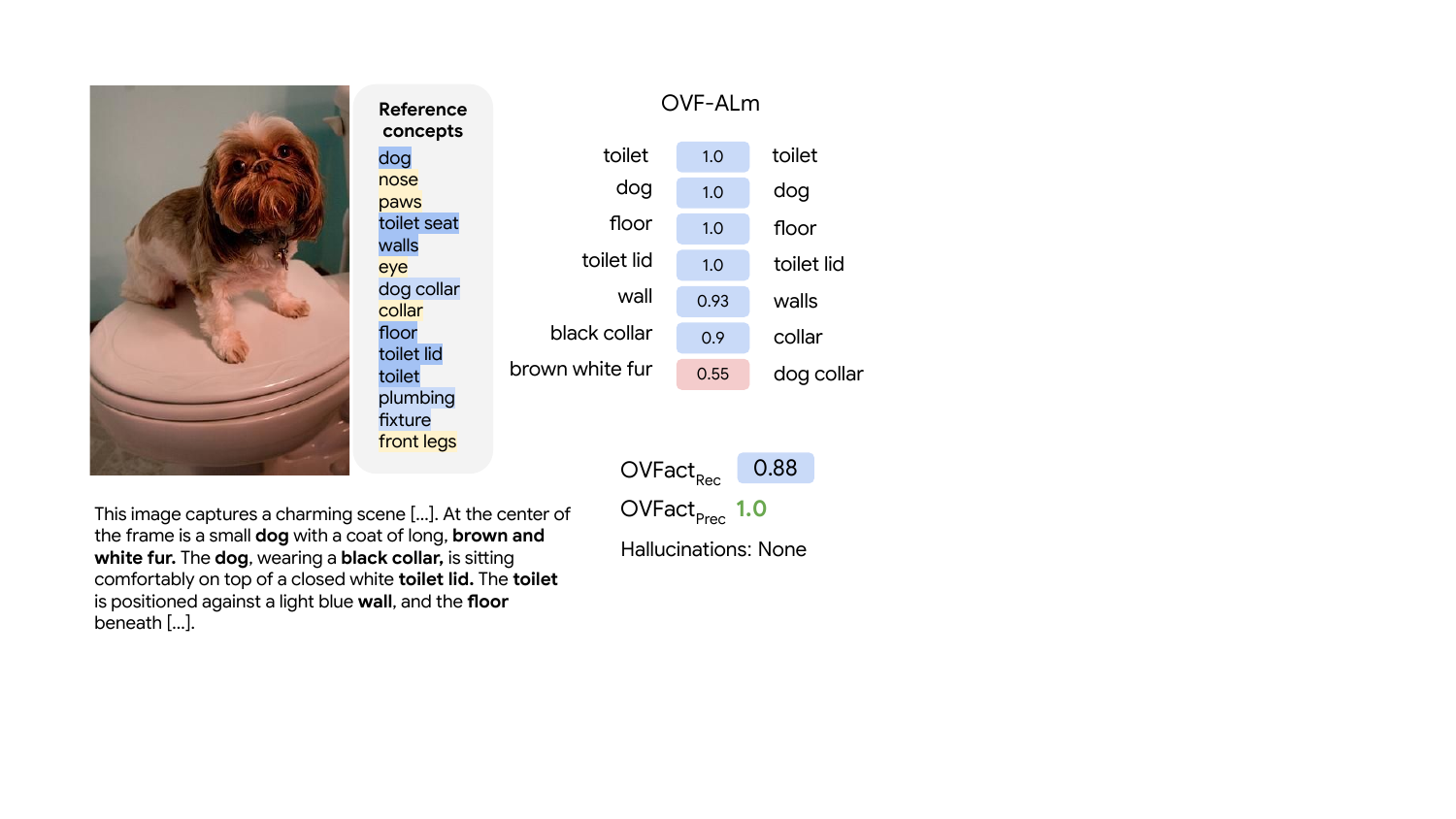}
    \caption{\textbf{Qualitative example of our filtering approach.} We present two examples from the ShareGPT4v dataset, showing how our method improves over our strongest baseline (our hybrid OVF-ALm method, which illustrates the brittle nature of the ALOHa matching method). For each example, we show the matching results for OVF-ALm to contextualize why this method incorrectly includes or does not include that example (Note that the OVF-ALm ``score'' is the minimum value of the set as with ALOHa, i.e., 0.78 on the left and 0.55 on the right). We also show the recall-precision breakdown of \ours{} score and the detected hallucinations by our method. \textbf{(left)} Example that was \textit{filtered out by our method} from the training mixture at a 0.4 data ratio but was kept in the baseline OVF-ALm (our strongest baseline, with ALOHa matching). Our method correctly identifies "tree" as a hallucination, while the compared baseline misses it. \textbf{(right)} Example with a clean caption within the 0.4 data ratio mixture of Ours, and is (incorrectly) not included in the baseline filtered set due to enforced one-to-one matching between "brown white fur" and "dog collar". See Sec.~\ref{app:qualitative_examples}.
    }
    \label{fig:app_qual_examples}
\end{figure*}
\para{Qualitative: Data filtering.} We present in Fig.~\ref{fig:app_qual_examples} qualitative examples of our method, applied to data filtering. In particular, we present two examples from the ShareGPT4v dataset, showing how our method improves over our strongest baseline (our hybrid OVF-ALm method, which illustrates the brittle nature of the ALOHa matching method). For each example, we show the matching results for OVF-ALm to contextualize why this method incorrectly includes or does not include that example. We also show the recall-precision breakdown of \ours{} score and the detected hallucinations by our method. The left image is an example that was \textit{filtered out by our method} from the training mixture at a 0.4 data ratio but was kept in the baseline OVF-ALm (our strongest baseline, with ALOHa matching). Our method correctly identifies "tree" as a hallucination, while the compared baseline misses it. The right image is an example with a clean caption within the 0.4 data ratio mixture of Ours, and is (incorrectly) not included in the baseline filtered set due to enforced one-to-one matching between "brown white fur" and "dog collar".

\begin{figure}[t!]
    \centering
    \includegraphics[width=\linewidth, trim={0em, 0em, 0em, 0em}, clip]{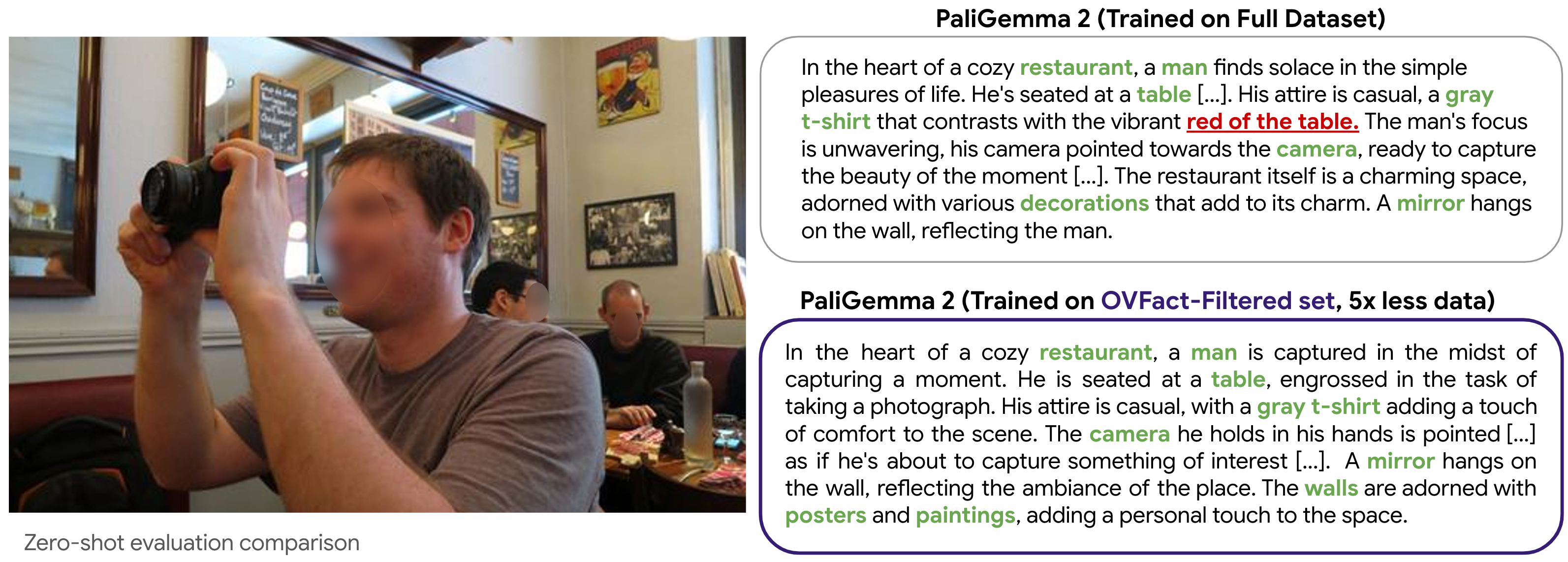}
    \vspace{3mm}
    \caption{\textbf{Qualitative example of trained model outputs.} We present an example comparison of two trained PaliGemma 2 models, \textbf{(top)} one on the \textit{full} training set, and \textbf{(bottom)} one on our \ours{}-filtered training set, with 5x less training data; both models are run zero-shot here on the Localized Narratives downstream dataset. We observe our model outputs consistently have lower rates of hallucination, without compromising descriptiveness. See Sec.~\ref{app:qualitative_examples}.
    }
    \label{fig:app_qual_examples_model}
\end{figure}

\para{Qualitative: Model outputs.} We also present in Fig.~\ref{fig:app_qual_examples_model} example outputs from two trained PaliGemma 2 models, one on the \textit{full} training set, and one on our \ours{}-filtered training set, with 5x less training data; i.e., a model trained on the full ShareGPT4v vs. our subset. Both models depicted are run zero-shot on the Localized Narratives downstream dataset. We observe that our qualitative model outputs (on both DOCCI and Localized Narratives) consistently have improved factuality precision, without compromising descriptiveness/recall. We also provide S$\times$S quantitative analysis with human evaluation in Sec.~\ref{app:traditional_metrics}.

\subsection{Models trained with \ours{}-filtered data: Traditional Metrics and Human Evaluations}
\label{app:traditional_metrics}
\label{app:human_filtering}
As mentioned in the main results section, our method improves factuality without compromising descriptiveness. In the main paper, we describe results across challenging long-caption downstream benchmarks and also show consistent results with prior/standard metrics (CHAIR) on specific domains (MS-COCO). 

\para{Traditional metrics.} Regarding traditional overall quality metrics, as noted by \citet{onoe2024docci}, metrics that may provide a signal for short captions (e.g., BLEU, CIDEr, etc.) do not provide a meaningful signal for long, descriptive captions when compared with human preferences. Thus, we do two things to evaluate models trained on our filtered subsets: (1) we report perplexity to ensure we are not sacrificing this for our other metrics gains, and (2) we provide human preference analysis of a model trained on our split vs. a model trained on the full dataset. Our gains in factuality reported in the main paper do not come at undue cost to perplexity (a traditional metric for assessing the overall diversity of the caption, more suitable for long captions) across both DOCCI and Localized Narratives.

\noindent Specifically, at the 20\% ratio (numbers below are (DOCCI, Localized Narratives) zero-shot evaluations):\\
{\small
\noindent \textbullet~Random filtering: ${\bf (2.605 \pm 0.004, 2.955 \pm 0.010)}$\\
\noindent \textbullet~Perplexity filtering: $(2.601 \pm 0.002 , 2.938 \pm 0.006)$\\
\noindent \textbullet~ALOHa filtering: $( 2.603 \pm 0.003 , 2.958 \pm 0.001 )$\\
\noindent \textbullet~OVF-ALm filtering: $(2.610 \pm 0.003 , 2.944 \pm 0.008 )$\\
\noindent \textbullet~Ours: $(2.610 \pm 0.003 , 2.956 \pm 0.001)$
}

\noindent We emphasize that in the above, our goal is to show that the perplexity for all models is effectively \textit{in the standard deviation window of the random filtering baseline} -- in other words, our metric does \textit{not} sacrifice perplexity to achieve all the gains we observe in factuality precision and recall. We also reiterate that developing strong metrics for long caption settings remains an open problem, one in which we hope \ours{} can play a supporting role. Finally, we note that even for the perplexity metric, the perplexity filtering does not seem to gain an advantage -- while prior work \cite{li-etal-2024-role} observed promising results applying perplexity/loss filtering, this was only in the setting of small-scale, short sentences with human-generated captions; in our setting, with model-generated noisy data at a large-scale, it seems this signal is not as clear and does not result in models that will then generalize well zero-shot to new downstream datasets.

\para{Human evaluation (Model outputs).} Finally, we validate our method on a stronger metric: human preference. We run a similar evaluation as we did with the metric analysis, but this time, we compare our model trained on our OVFact-filtered data (5x less than the full set) against a model trained on the full set. 
We use images from DOCCI \textit{qual test} dataset and ask 3 independent annotators to give their preference. In this side-by-side comparison, model outputs for the same image are compared, and the order of model A (left) and model B (right) in the UI is randomized to prevent order bias and we repeat this for the DOCCI evaluation set with at least two human evaluations per image like before. This time, we care about model preference, and the question posed to evaluators asked about general quality with both descriptiveness and accuracy. We observe that humans select our model over the model trained on the full dataset with a strong preference rate of \textbf{68.2\%}. This indicates that our metric improvements correlate with overall quality improvements and that models trained on our filtered subset have higher quality with 5x less training data.

\subsection{Additional experimental details}
\label{app:experimental_details}

We implement our approach in the Big Vision~\cite{big_vision} codebase in JAX~\cite{jax2018github} and choose models for our experiments and base model tools with non-restrictive open-source licenses (all model use in accordance with intended usage). Our implementation details consist of two key aspects: (1) the model training for the data filtering experiments and (2) our open-source models (tools) that are used for our metric. We elaborate on key details below and plan to release further supporting material to facilitate the adoption of our method in the broader community.

We train all our PaliGemma 2 \cite{steiner2024paligemma} models with input images of size 448 x 448, following the best practices outlined in the original paper. Our main experiments in the paper focus on the PaliGemma 2 model with 3B parameters (SigLIP-So400m encoder with 14x14 pixel patches, yielding 1024 tokens per image, and a 2B parameter LLM decoder), but we show results for other model sizes in Section~\ref{app:model_study}. We train our model with a batch size of 128 (image, caption) examples for 5 epochs over the full input large-scale training set. Following \cite{steiner2024paligemma}, we use the Adam optimizer \cite{kingma2014adam} with default hyperparameters (b2 = 0.999, grad clip norm = 1.0, etc.) throughout and adjust the learning rate for the different model sizes (e.g., learning rate 1e-6 for 3B). The model card for PaliGemma 2, detailing its pre-training data mixture, is available through the official model release on github and Huggingface, which can be found in the paper \cite{steiner2024paligemma}.

For our tools, as discussed in Sec.~\ref{sec4:exp_setup}, we use state-of-the-art open-source models to ensure the broader community can also run our method with consistent results. We leverage for our LLM Gemma2-27b~\cite{team2024gemma} for caption parsing (see Sec.~\ref{app:llm_prompt} for prompt), and OWL-ViTv2~\cite{minderer2024scaling} and OpenSeg~\cite{openseg} for entity grounding (open-vocabulary detection and segmentation). As per Sec.~ \ref{sec:3_approach}, to assess descriptiveness (recall) in a reference-free manner, we consider an extensive vocabulary $\mathcal{V}$ of open-vocabulary concepts. Specifically, we take the union of concepts from standard datasets, i.e. Visual Genome \cite{krishna2017visual}, LVIS \cite{gupta2019lvis}, Open Images~\cite{kuznetsova2020open} and Objects365~\cite{shao2019objects365} resulting in a total of 2792 unique concepts. To encode entities for \avgsim{} computation, we use SigLIP-So400m/14~\cite{zhai2023sigmoid} text encoder.
We also highlight that our \ours{} is intended to be a general framework that can continue to incorporate other base models as these continue to improve. 

\subsection{LLM prompt}
\label{app:llm_prompt}
We present the LLM prompt we use for caption parsing in Fig.~\ref{fig:app_llm_prompt}. As discussed in the main paper, our goal for the model is to ensure that it captures key, \textit{groundable} objects from the input text $y$ (VLM / model caption) and to output a final candidate set $\mathcal{C}$ for later use in our method.

\begin{figure}[t]
    \centering
    \includegraphics[width=0.99\linewidth, trim={1em, 0em, 1em, 3em}, clip]{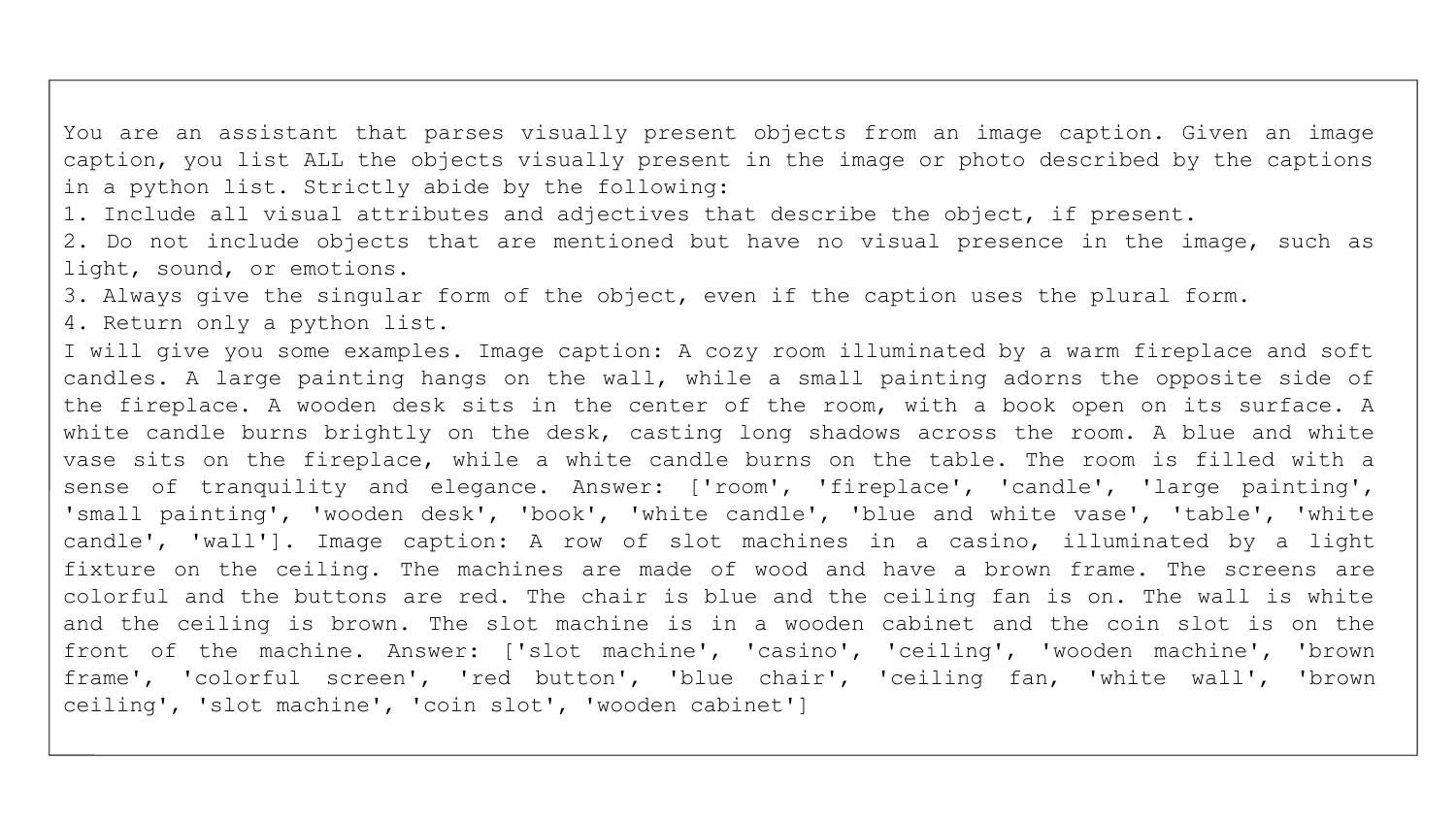}
    \vspace{-10mm}
    \caption{\textbf{Gemma 2 prompt.} We provide the prompt details above for the Gemma 2 parsing step, which maps the input text $y$ to the candidate set $\mathcal{C}$.}
    \label{fig:app_llm_prompt}
\end{figure}

\begin{figure}[t!]
    \centering
    \includegraphics[width=\linewidth, trim={0em, 0em, 0em, 0em}, clip]{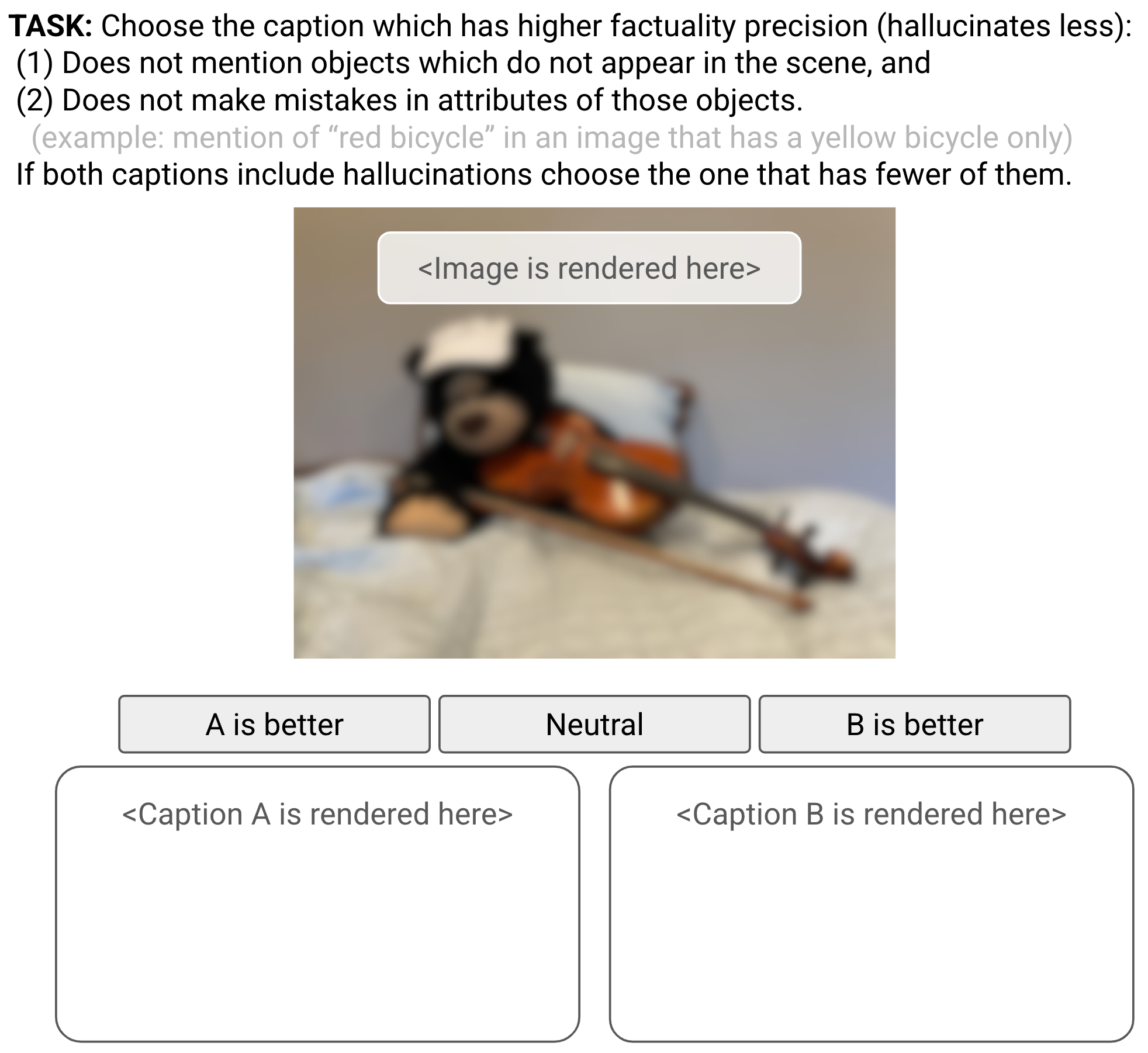}
    \caption{\textbf{UI interface example.} We show a screenshot of our UI for our human studies of S$\times$S comparisons (blurring the image and removing the captions). We extend the analysis from \citep{onoe2024docci}. See Sec.~\ref{app:human_alignment} and Sec.~\ref{app:traditional_metrics} for discussions of human evaluations of our method: we show consistent improvements over key baselines, for both measuring agreement, and better quality model outputs for models trained on our filtered subset.
    }
    \label{fig:app_qual_examples_model}
\end{figure}

\subsection{Discussion on efficiency}

\begin{table*}[t]
\small
\centering
\begin{tabular}{l|ccccc}
\toprule
Number of text concepts & 1& 30 &	300	& 2,792 (Ours)	& 300,000 (>100x Ours) \\
\midrule
Runtime in \textit{miliseconds} &  7.35 $\pm$ 0.12  &  7.54 $\pm$ 0.13	  &  7.57 $\pm$  0.32 & 7.58 $\pm$
 0.24 & 12.93 $\pm$ 0.26 \\
\bottomrule
\end{tabular}
\caption{\textbf{Runtime scaling experiments of OWL-ViT w.r.t. number of prompted text queries (concepts).} We validate that our method scales well with increased number of concepts $V$.}
\label{app:tab_efficienc}
\end{table*}

We design our method with efficiency and scalability in mind since we explore a novel application of our method to large-scale pretraining datasets.

We note that the grounding tools (OWL-ViT, OpenSeg) we chose in this work have a “late fusion” architecture, which means that they are extremely scalable in terms of the number of text queries. This is because the input image and text queries can be embedded separately, text embeddings are only used at the “end” of the network (the bulk of visual processing is query-agnostic), and text embeddings can be cached and reused across different inference calls. To illustrate by example, we focus on our OWL-ViT model in detail below.

This is confirmed by the table below where increasing the number of text queries from 1 to 2792 (the total size of our reference set), increases the runtime by only 3\% (< 1 millisecond).
Note that the timing was performed on a NVIDIA A100 GPU using an image resolution of 448 x 448. While we precomputed text embeddings for the above table, we note that it only takes 0.84 ms to compute all 2792 text embeddings, which is an order of magnitude less than the rest of the network (and which needs to be only done once).

\end{document}